\documentclass[review]{elsarticle}
\journal{Computers and Operations Research}
\usepackage[colorlinks=true, urlcolor=blue, pdfborder={0 0 0}]{hyperref}
\usepackage{natbib}
\biboptions{authoryear}
\usepackage[margin=2.5cm]{geometry}
\usepackage{array}
\usepackage{changepage}
\usepackage[ruled,vlined,linesnumbered]{algorithm2e}
\usepackage[numbered]{bookmark}
\usepackage{amsmath,amssymb,amsthm}
\usepackage{mathtools}
\allowdisplaybreaks[1]
\usepackage{tikz}
\usepackage{ulem}
\usepackage{longtable}
\usepackage{soul}
\usepackage{xfrac}
\usepackage{color, colortbl}
\usepackage{siunitx}
\usepackage{placeins}
\usepackage{textcomp}
\usepackage{bbding}
\usepackage{rotating}
\usepackage{multirow}
\usepackage{booktabs,siunitx}
\usepackage{array}
\usepackage{graphicx}
\usepackage{animate}
\usepackage{tikz}
\usepackage{lipsum}
\usepackage{float}
\usepackage{blindtext}
\usepackage{subcaption}
\usepackage{parskip}
\usepackage{soul}

\usepackage[margin=2.5cm]{geometry}
\usepackage{array}
\usepackage{changepage}
\usepackage[numbered]{bookmark}
\usepackage{mathtools}

\usepackage{ulem}
\usepackage{soul}
\usepackage{xfrac}
\usepackage{color, colortbl}
\usepackage{placeins}
\usepackage{textcomp}
\usepackage{bbding}
\usepackage{rotating}
\usepackage{multirow}
\usepackage{booktabs,siunitx}
\usepackage{animate}
\usepackage{algorithmicx}
\usepackage[noend]{algpseudocode}
\usepackage[thinc]{esdiff}
\usepackage[T1]{fontenc}

\DeclareSIUnit \year {year}
\DeclareSIUnit \kWh {kWh}
\DeclareSIUnit \tons {tons}
\DeclareSIUnit \rev {rev}
\renewcommand\appendix{\par
  \setcounter{section}{0}
  \setcounter{subsection}{0}
  \setcounter{figure}{0}
  \setcounter{table}{0}
  \renewcommand\thesection{Appendix \Alph{section}}
  \renewcommand\thefigure{\Alph{section}\arabic{figure}}
  \renewcommand\thetable{\Alph{section}\arabic{table}}
}
\theoremstyle{plain}
\theoremstyle{plain}
\theoremstyle{plain}
\begin{document}
\begin{frontmatter}

\title{How to effectively use machine learning models to predict the solutions for optimization problems: lessons from loss function}

\author[Mahdiaaddress]{Mahdi Abolghasemi\href{https://orcid.org/0000-0003-3924-7695}{\includegraphics[scale=.6]{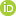}} }\ead{mahdi.abolghasemi@monash.edu}
\author[myaddress]{Babak Abbasi\href{https://orcid.org/0000-0002-4332-2662}{\includegraphics[scale=.6]{1.png}}} \corref{mycorrespondingauthor}
\cortext[mycorrespondingauthor]{Corresponding author}
\ead{babak.Abbasi@rmit.edu.au}
\author[Toktamaddress]{Toktam Babaei}
\ead{toktam.babaei@gmail.com }
\author[Zahraaddress]{Zahra HosseiniFard
\href{https://orcid.org/0000-0002-1064-9079}{\includegraphics[scale=0.6]{1.png}}}
\ead{zahra.h@unimelb.edu.au}

\address[Mahdiaaddress]{Department of Econometrics and Business Statistics, Monash University, Caulfield, VIC 3162, Australia}
\address[myaddress]{Department of Information Systems and Business Analytics, RMIT University, Melbourne, VIC 3000, Australia}
\address[Toktamaddress]{School of Mathematical Sciences, Queensland University of Technology, Brisbane, QLD 4000, Australia}
\address[Zahraaddress]{Faculty of Business and Economics, The University of Melbourne, Parkville, VIC 3010, Australia}

\begin{abstract}

Using machine learning in solving constraint optimization and combinatorial problems is becoming an  active research area in both computer science and operations research communities. This paper aims to predict a good solution for constraint optimization problems using advanced machine learning techniques. It extends the work of \cite{abbasi2020predicting} to use machine learning models for predicting the solution of large-scaled stochastic optimization models by examining more advanced algorithms and various costs associated with the predicted values of decision variables. It also investigates the importance of loss function and error criterion in machine learning models where they are used for predicting solutions of optimization problems. We use a blood transshipment problem as the case study. The results for the case study show that LightGBM provides promising solutions and outperforms other machine learning models used by \cite{abbasi2020predicting} specially when mean absolute deviation criterion is used. 
 
 \medskip

\end{abstract}


\end{frontmatter}

\section{Introduction} \label{sec:intro}

Many organizations are faced with complex decision problems in their day-to-day operations that can be formalized as a stochastic optimization problem. For example, in blood inventory management systems, hospitals have to make ordering decisions from central blood bank as well as transshipment decisions in a network of hospitals under uncertain demand and perishable nature of blood products. In many cases, the solutions to these problems are required instantly, while, optimal solutions might only be available through commercial optimization solvers with a considerable amount of time for a large-scale decision problem. Due to these barriers, implementation of optimization techniques in some industries remains a challenge.
In this paper, we investigate applications of advanced machine learning (ML) techniques for solving large-scale stochastic optimization problems through predicting the solutions. Further, we investigate the role of different loss functions in predicting the optimal solutions for such a problem. We analyze various loss functions including mean absolute error (MAE), mean squared error (MSE), and Huber loss in optimizing the predictive machine learning models for effective learning from data.

ML models are widely applied in various areas including automation of operations in supply chain and warehousing, forecasting models, customer segmentation, image processing, speech recognition and so on. A supervised ML technique can predict the output associated with new inputs through learning from input-output mappings. 
From the operations research literature, ML techniques are mainly developed as a heuristic solution to an optimization problem or to improve the accuracy of solver algorithms (e.g., \cite{smith2009cross, vaclavik2018accelerating, kruber2017learning, lodi2017learning}). 
However, predicting the optimal solutions via ML models (i.e., from an exact solver) has not been explored enough. 
\cite{larsen2021predicting} studied prediction of tactical decisions using ML models, however, they did not determine the value of operational decision variables. \cite{abbasi2020predicting} proposed to predict the solutions for an optimization problem and the optimal value of decision variables using four machine learning models including classification and regression tree, k-nearest neighbors, random forest, and multilayer perceptron (MLP) from artificial neural network family. However, they did not consider the impact of loss function criterion in optimizing the ML learning algorithm. Loss function is an important criterion for effective learning from data and extrapolation and can significantly impact the generated solutions with ML models. We investigate the performance of different loss functions in optimizing the learning process of ML models for the  optimizations tasks. 

%Loss functions...
We build our study on the findings of \cite{abbasi2020predicting}. We further extend their work in three main directions. First, we explore the impact of loss function when predicting the solutions of optimization problems. We consider three ML algorithms, commonly used in forecasting tasks, namely Light Gradient Boosting Machine (LightGBM), Support Vector Regression (SVR), and Ridge regression to predict the optimal values of the decision variables. We train the LightGBM model with MSE, MAE and Huber loss to examine the performance of the loss function. Second, we further explore the performance of the models by examining the number of times proposed solutions have violated a constraint. We show that the loss function can significantly impact the solutions of the models and whether they violate any constraint or not. Third, we look at the forecast utility by evaluating the first stage objective function of a two-stage stochastic optimization problem (2SSP) for a blood transshipment case study. We evaluate its efficiency with various incurred costs including ordering, holding, transshipment, outdate and shortage costs.

The key findings of this study are as follows:
\begin{itemize}
    \item  By using finely tuned advanced machine learning models, we can predict the optimal values of the optimization problems and achieve high accuracy (up to 98\% similarity to the optimal policy) while committing to the constraints (over 99\% of the periods). Compared to commercial solvers, ML models can generate competitive results in a shorter amount of time with significant savings in cost. 
    \item Loss function plays a pivotal role in predicting the optimal solutions of the constrained optimization problems. 
    \item Looking at the utility of forecasts associated with the ML predictive models, we observe that ML models can outperform the optimal solution by reducing some other costs of the supply chain while being competitive in total costs.
\end{itemize}

The rest of the paper is organized as follows: Section \ref{sec:literaturereview} is a review of related literature. In Section \ref{sec:methodology}, we describe the methodology used throughout this study, and we present the proposed machine learning prediction algorithm. Data and experimental setup are explained in Section ~\ref{sec:setupdata}, while Section \ref{sec:MLresults} provides the empirical results and some discussions of findings. Finally, we conclude paper and provide some future research directions in Section ~\ref{sec:MLconclusion}.

%Forecasting is the basic for many of the managerial decisions that involve uncertainty. It has been shown that accurate forecasting of demand can increase service levels \citep{pooya2019exact}, decreases the inventory costs \citep{syntetos2006stock} and in overall increase the performance of the supply chain \citep{syntetos2016supply,zamani2020considering}. 

%The remainder of the paper is organised as follows. Section \ref{sec:literaturereview} critically reviews the literature. Section~\ref{sec:methodology} introduces our methodology. Section~\ref{sec:setupdata} presents the dataset used for the empirical evaluation of the proposed approach and describes the experimental setup. Section~\ref{sec:MLresults} presents the results of the experiment and discusses our findings. Finally, section~\ref{sec:MLconclusion} concludes the paper.

\section{Literature review} \label{sec:literaturereview}

%\textcolor{violet}{
Two-stage stochastic programming was introduced by \citet{dantzig1955SSP} to tackle uncertainty problem in mathematical programming. This approach has been applied in inventory management systems since then. 
One of the applications of stochastic optimization models is in blood inventory management. %}
Given the limitation of blood donor population, blood inventory management has become a critical task to avoid the fatal risk of insufficient number of blood products in hospital inventory. In addition to the uncertain nature of supply and perishable characteristic of blood, the demand for blood products is also uncertain.
% \textcolor{violet}{\st{If the demand were constant, or known with certainty well in advance, then the operation of a supply chain would be a straightforward (backwards) scheduling exercise. However, demand is not known and the uncertainty associated with blood demand makes blood supply chain management a challenging and important matter}}.
Due to these complexities, heuristic-driven solutions become the common practice in blood inventory management while it may not provide the optimal solution \citep{Dehghani2018}. In order to achieve an optimal solution, some researchers utilised combinatorial optimization models such as linear and mixed-integer linear programming to obtain exact solutions of the problem. %\textcolor{violet}{reference??}.
\citet{DehghaniTwoStage} proposed a 2SSP framework for optimizing a blood supply chain consisting of four hospitals and a central blood bank and included a proactive transshipment policy in their model. They compared the costs associated with their optimized ordering and transshipment policies with a no-transshipment policy and showed the superiority of their model. However, solving the blood transshipment stochastic optimization problem involves heavy computations and accessing to somehow expensive commercial software tools. 

Alleviating the burden of computational time on one hand and the abundance of available data from supply chain systems, on the other hand, motivated researchers to use ML techniques to predict solutions \citep{larsen2021predicting,abbasi2020predicting, mossina2019multi}. 
% \textcolor{violet}{
%which is the motivate of this paper as well. In this research, we investigate the results of using three advanced ML models including LightGBM, Support Vector Regression, and Ridge regression in prediction the optimal solution with different loss functions. We use the same case study as \cite{abbasi2020predicting}'s paper for the comparison purposes.
ML is a natural approach to perform on problems that include data with the unknown exact analytical distribution. Applications of ML in combinatorial optimization problems can be generally divided into two streams, those who used ML to mitigate the computational burden associated with mathematical method, and those who used ML to solve problems which are not mathematically well defined \citep{bengio2020MLsurvey}. While the latter case builds a solution from scratch, the first approach is a supervised learning task. In a supervised learning task, the algorithm learns the solution by looking at examples solved by an expert (known as training data). A trained algorithm is evaluated by its generalization ability, i.e., its performance over new unseen situations (known as test data). The common approach is to collect training data offline and then present it to model. However, a few recent methods developed advanced algorithms which are able to receive training data in an online manner and increase robustness and stability of model \citep{marcos2016online}.
\citet{abbasi2020predicting} formulated the aforementioned blood supply chain optimization problem into a supervised ML framework and used the solutions of 2SSP models as training data set to build models. More specifically inventories of hospitals at each day were the inputs and decisions on the number of orders and transshipments were considered as output variables. They evaluated the performance of most common ML models in terms of practical utilities by simulating such blood supply chain over a period of 18500 days. This study aims to expand their work by using different and advanced ML models that are able to perform better both in terms of accuracy (not violating constraints) and efficiency (lower costs). We specifically elaborate on the importance of loss function in addressing optimization problems with ML, and empirically investigate the performance of different loss functions in learning and predicting the values of decision variables.

%Inventories of the hospitals considered as inputs while decisions on number of orders and transshipments.   
%However, there are no clear guidelines on the applying ML to the problem. 

With the growing complexity of data, it has become important to optimize the ecosystem of the ML models such as pre-processing, the number of parameters, and finding the optimal values of hyper-parameters to enable models to learn effectively from data and extrapolate the future. One such important criterion is the loss function. The loss function is the way that a predictive model learns the relationship between inputs and outputs, thus the learning process differs if we change the loss function of the model. %Consequently, the in-sample accuracy of the model changes which may impact the out-sample accuracy of our predictive model.   
Forecasting models, either statistical or ML, can be optimized by using a loss function. 
We can choose either a linear loss function such as MAE, quadratic such as MSE, cubic or higher orders depending on the problem at hand. 
There have been several studies exploring the impact of loss function on the performance of the forecasting models. Some studies have tried to find the optimal loss function for their problem and data set. For example, \cite{barron2019general} found that in neural network architectures, $L^2$ loss is effective for models with multivariate outputs that depict the different level of robustness across its dimensions. In classification tasks, it has been shown that variations of the cross-entropy function have improved the performance for certain types of classification tasks \citep{lin2017focal}.
\citet{makridakis1991exponential} investigated the impact of different loss functions on the post-sample accuracy of various exponential smoothing models. They applied MAE, MSE and MAPE on 1001 time series and measured the post-sample accuracy with MAE, MAPE, and MSE. They concluded that when the median is used to optimize the parameters, there were inconsistencies in the accuracy of models regardless of the evaluation metric, i.e., MAE, MAPE, MSE. They did not find any pattern when they employed MAE, MAPE and MSE to optimize the model's parameters and measure their performance. They suggested using MSE as a more robust loss function. Linear errors such as MAE penalise overestimation and underestimation equally. However, in reality, this may not be the case as often underestimation is more costly. Therefore, the asymmetric loss function has been put forward to train and evaluate the forecasting models \cite{christoffersen1997optimal}.

There is neither a unique loss function that works well on all data set and for all problems nor does it exist a forecasting model that works well on all sorts of data. There are indeed horses for courses and forecasters need to find an appropriate model and loss function for their problem \citep{Petropoulos2014a}. While there are some general guidelines to choose the appropriate type of models and loss function, researchers are more reliant on experiments to find their desirable features \citep{abolghasemi2020demand,makridakis1991exponential,abolghasemi2020demand2,zamani2020considering}. With the advancement of ML models and their capability to have customised loss function, there has been a growing interest in developing custom loss function for various predictive modelling tasks \citep{MONTEROMANSO202086,lin2017focal, barron2019general}. 
\citet{MONTEROMANSO202086} used a custom loss function to minimise the forecasting loss aroused from selecting forecasting models. They adopted the loss function of an extreme gradient boosting model to use a weighted mean absolute error loss for minimising the error of combining the forecasts generated from a pool of methods. In fact, they used a customised loss function to find out the optimal weights for various forecasting models to minimise the out-of-sample errors. Their methods ranked second in the well-known M4 forecasting competition owing to the careful selection of models by using a customised loss function.

%In evaluating forecasting models, literature have emphasised on the differentiation between forecast accuracy and forecast utility. 

Another important phenomenon in forecasting, and in particular, supply chain forecasting problems is the utility of forecasts. The common approach for evaluating the performance of forecasting tasks is to compare the accuracy of the generated forecasts against their actual values. However, in business forecasting applications, statistical measures do not necessarily reflect their actual performance. In other words, managers are often interested in the profit or loss resulting from their forecasting models, not the accuracy of the forecasting models \citep{armstrong2001principles,clements1995selection}. 
 There has been a long-standing debate among researchers with regards to the suitability of popular loss functions such as MAE and MAPE in supply chain setting. While these measures might be relevant for some problems, they are criticized for their bias and irrelevancy in various problems \citep{boylan2006accuracy,syntetos2016supply}. Therefore, utility measures such as total inventory costs and production costs have been put forward as a more suitable way of measuring the performance of forecasting models \citep{ali2012forecast}. The utility of forecast has been minimally considered with the majority of researchers looking at the utility of forecasts only on inventory costs \citep{syntetos2006stock}.  In this study, we take a holistic approach for monitoring the utility of forecasts in the selected blood supply chain case study and by examining the ordering, transportation, shortage, outdate, and holding costs imposed by the generated forecasts. 
 %That is we look at the utility of forecasts by projecting the impact of the predicted decision variables in holding (inventory), ordering, transportation, outdate (wastage) and shortage costs.

We summarise our contribution to the previous studies as follows:
\begin{enumerate}
    \item We propose to use an appropriate type of loss function to train the ML models, in particular when using them to emulate the solutions of constrained optimization problems. We implement LightGBM models with three different loss functions along with Ridge regression and SVR models to predict the solutions of stochastic optimization problems.
    \item We show that by finely tuning the ML models and using appropriate loss function we can minimise the number of constraint violations aroused from using ML models to predict the solutions of constrained optimization problems. 
    %We look at the properties of data and investigate the role of various loss functions in predicting the solutions of mathematical problems. 
    \item We look at the utility of forecasts associated with the predictions of ML models. We extract the holding, ordering, outdating, shortage, and transportation costs in an empirical supply chain case study problem and show that the proposed ML models can outperform the optimal policy in some of the incurred costs.
\end{enumerate}

\section{Methodology} \label{sec:methodology}

We posit that we can use an ML model to learn the relationship between a set of input and output decision variables on the first stage of a 2SSP problem. Although the predicted solution may not be optimal, it has some advantages over the mathematical models. Firstly, if the ML model can learn the input-output mapping without loss of accuracy, decision makers can use that in a matter of second as opposed to mathematical models that may require a few minutes to solve. Secondly, ML models can replace the commercial software which are often costly both in terms of licence and human labors.

%While the number of first stage decision variables are often limited, the number of second stage decision variables can increase dramatically as the number of scenarios and the length of planning horizon increases. 

%to determine the values of decision variables.
%Note that, we use ML model to generate solutions only for the first stage of the objective function of the blood supply chain model shown in \ref{optimizationmodel}. 
In our case study, the data used for training the ML models consists of 18,500 observations with 44 inputs and 136 outputs. The input variables include the inventory levels of 11 different blood units with different ages for each hospital totalling 44 variables.  The output variables correspond to the transshipment of different aged blood units between hospitals, and orders of hospitals, totalling 136 decision variables.  In order to train and test the performance of the ML models, we split the original dataset into train and test sets. We used the first 16,650 observations of data to train the ML models. We then use the trained models with inventory levels as inputs to predict the blood transshipments between hospitals.

%A two-stage stochastic optimization in operational decision making, considering $S$ as the planning horizon, was formulated by \cite{DehghaniTwoStage}.

The ML models use the inputs and solutions of the 2SSP model developed by \cite{DehghaniTwoStage} as the training data in a multi-input multi-output fashion to predict the decision variables.
The proposed methodology is summarised in the algorithm \ref{MLSCMAlgorithm}. The original algorithm is derived from \citep{abbasi2020predicting} and adjusted to reflect the choice of loss function and compute various costs in here.\\
\vspace{4mm}
\begin{algorithm}[H]
	\begin{algorithmic}[1]
		\State 	\textbf{Off-line phase}
		\State Formulate the problem in hand with an appropriate mathematical model. 
		\State Solve the mathematical model with a numerical solver to find the optimal solution.
		\State Build a simulation model that mimics the supply chain behaviour in determining the values of decision variables.
		\State Run the simulation model to obtain the decisions according to the mathematical model developed at step 1, then store the parameters used as inputs and the decision variables as outputs for training the ML models. 
		\State 	\textbf{On-line phase (Rolling horizon with length of $S$)}
	    \State Split the the dataset obtained in step 3 into training and test sets. 
	    \State Develop appropriate ML models with an appropriate loss functions and train them using the training dataset.
	    	\State \textbf{For {$t =1 ~to ~ S $ }}
        	\State Predict the decision variables on a rolling origin basis, and replace the obtained values of the decision variables in the simulation model.   
        	\State Evaluate the predictions by replacing the obtained values in the objective function of the mathematical model and computing various costs of each decision that has been made. 
			\State \textbf{End For} 
			%\caption*{ML prediction algorithm} 	\label{MLSCMAlgorithm}
	\end{algorithmic}
	\caption{ML prediction algorithm} 	\label{MLSCMAlgorithm}
\end{algorithm}	

\vspace{4mm}
With regards to the choice of ML model, we admit that there is no unique model that is capable of forecasting all types of data more accurately than other models under all conditions \citep{abolghasemi2020demand}. However, various empirical studies suggest some effective models for certain types of data \citep{FILDES20151692,Petropoulos2014a}. Given that we have a sparse data comprising the parameters of the optimization model as input and the optimal values of decision variables as outputs, we use ML models with regularization. Note that the algorithm \ref{MLSCMAlgorithm} is flexible in terms of the method that can be employed for predicting the solutions. That is, users can choose their ML regression method of choice using this algorithm for predicting the decision variables of the optimization model.  We implement three ML algorithms, including Ridge regression, SVR, and LightGBM with various loss functions that are shown to perform well in various forecasting tasks \citep{abolghasemi2020demand,abolghasemi2019machine,ke2017lightgbm,drucker1997support,hoerl1970ridge}.

\subsection{Ridge regression}
Ridge regression is a powerful regression model that deals with multicollinearity problem in linear regression \citep{hoerl1970ridge}. The multicollinearity often occurs when the number of parameters are large and potentially they are not independent. Ridge regression performs $l2$ regularization technique with a penalty cost, $\lambda$ to estimate the parameters of the model and avoid unbiased results. That is, it adds an identity matrix as noise to the cross product matrix in order to obtain a reliable estimation for the parameters. The ridge loss function is defined as follows in equation \ref{ridgeReg}: 

\begin{align}\label{ridgeReg}
    L (\beta, \lambda)= \sum_{i=1}^{n}(y_i -\hat{y_i})^2{n} + \lambda\sum_{j=1}^{p} \beta^2, 
\end{align}

where $y$ is the actual value of the target variable and $\hat{y}$ is the predicted value of the target variable, $\lambda$ is the ridge estimator, $n$ is the number of observations and $p$ is the number of variables. The first part of the loss function is simply the sum of squared error loss function and the second part is the ridge penalty.  
For a convex function, the ridge loss function guarantees that it rests at the global minimum by finding the $\beta$ and $\lambda$ values. 
Since the range of parameter values in the output vector changes significantly,  the ridge regression penalises the large model weights making it suitable for obtaining unbiased results for blood supply chain model. This model has been successfully implemented in various forecasting applications \citep{exterkate2016nonlinear}. We use this model in multi-input multi-output fashion and apply a 10-fold cross-validation technique to obtain the optimal value of $\lambda$.

\subsection{SVR}
SVR is a powerful supervised learning algorithm \citep{burges1998tutorial}. SVR is different from an ordinary least square regression model in the sense that SVR attempts to minimise the generalised error rather than minimising the deviation of predicted values from the actual ones. SVR maps the input data to a higher dimensional space in a non-linear fashion using a kernel function. SVR finds a function in the space within a distance of $\epsilon$ and $\epsilon^*$ from its predicted values. Any violation of this distance is penalised by a constant penalty cost,$c$. For a given set of data points ($x$, $y$), SVR solves the following constrained optimization problem to estimate the parameters. 

\begin{align}
& \min_{w,b,\xi,\xi^*} \frac{1}{2} w^Tw + C \sum_{i=1}^n \xi_i + C \sum_{i=1}^n \xi_i^* \\
& y_i - w^T \phi(x_i) - b \leq  \epsilon + \xi_i ,  \\
& w^T \phi(x_i) + b -y_i\leq  \epsilon + \xi_i^* ,  \\
& \xi_i, \xi_i^* \geq 0, i = 1, \ldots, n. 
\end{align}\label{svr}

SVR has been effectively applied in many different forecasting applications including supply chain forecasting problems \citep{levis2005customer,abolghasemi2020demand2,abolghasemi2020demand}. We implement SVR in a multi-input multi-output fashion. We chose the Radial Basis as the kernel function and optimize the cost of the constraint violation, $C$, using a 10-fold cross-validation technique. The minimization of the error rate was used as a loss function.

\subsection{LightGBM}
LightGBM is an implementation of gradient boosted decision trees that is based on ensembling and uses a number of hyperparameters for training models and generating predictions \citep{ke2017lightgbm}. LightGBM has been implemented in various forecasting problems and attracted many researchers and practitioners attention by winning a number of forecasting competitions including the latest M5 forecasting competition \citep{bojer2020kaggle,makridakis2021m5}. LightGBM is a fast and powerful algorithm that can handle various types of features, making it appealing for large scale problems with a large number of diverse input variables. We implemented this model in multi-output fashion to predict the solutions of the constrained optimization model. LightGBM algorithm benefits from a large number of hyperparameters to learn the process and project them to the gradient space. These parameters play a critical role in the performance of the model. We can find the optimal values of these parameters using 10-fold cross-validation technique. That is we divide the data set into 10 folds, where we iteratively use nine parts of data to train the model and one to test the performance of the model. The optimal value of the main hyperparameters are set as follows:
the learning rate (\texttt{eta}) to 0.01, \texttt{colsample-bytree} to 1, \texttt{min-child weight} to 5 , \texttt{max-depth} to 15, \texttt{sub sample} size to 0.7, and the number of iterations to 1000. We trained LightGBM by setting the objective to \textit{regression} and evaluated its performance with \textit{rmse}. In order to avoid overfitting issue which is a common problem in tree-based models, we use both the $L1$ and $L2$ regularization. 
One prominent feature of LightGBM model is its ability to train based on a different loss function that is desirable for decision-makers. 
We explain about loss function in the next part.

\subsection{Loss function}
Loss function is an important part of any learning process and forecasting problem. The true loss function is often difficult to estimate as its distribution is unknown. There is no consensus in the literature in choosing the best loss function \citep{clements1995selection,clements1993limitations}, rather it depends on the dataset, problem at hand, and the objectives of the decision-maker. While in academia statistical loss functions are widely used for training and evaluating models performances, the loss function in real world is measured in dollar terms for managers. Nevertheless, reliability, robustness to outliers and comprehensibility are some of the desirable criteria for a good loss function. 
We employ mean squared error (MSE), mean absolute error (MAE), and Huber loss as three widely-known loss functions and empirically evaluate their performance in blood supply chain problem.
We first use MSE as the loss function. MSE is widely used in regression problems and it is the default loss function for various ML and statistical models. MSE tries to find the best values of parameters by minimising the average error across all observations. The MSE loss is calculated as follows: 

\begin{align}
    MSE= \frac{1}{n}\sum_{i=1}^{n}\frac{(y_i -\hat{y_i})^2}{n}, 
\end{align}
where  $y$ is the actual value of the target variable and $\hat{y}$ is the predicted value of the target variable. 

For the second attempt, we used MAE as the loss function. MAE is another popular loss function that is frequently used in various settings. MAE optimize the learning process by considering median of the values. MAE loss is calculated as follows: 
\begin{align}
    MAE= \frac{1}{n}\sum_{i=1}^{n}|y_i - \hat{y_i}|,
\end{align}
where  $y$ is the actual target variable and $\hat{y}$ is the predicted value of the target variable. 
Since MAE tries to minimise the errors by considering the median of observations, it is an appropriate choice to avoid the large impact of outliers that may be imposed on the learning process. This is in contrast to MSE that optimize the mean across all predictions. MSE penalises any violation with a large cost making it a non-robust cost function especially when there are outliers in the predictions. While MAE is a robust estimator, it can be biased because the gradient is not dependent on the size of the error but only on the sign of the error. That is if the error is negative the gradient is -1, and when the error is positive the gradient takes the value of +1. This can be problematic and cause convergence problem when the error is small. 

We also implemented Huber as another well-known loss function. 
Huber loss is calculated as follows: 
\begin{align}
    Huber_{\delta}(y, \hat{y}) = 
 \begin{cases} 
  \frac{(y - \hat{y})^2}{2}, \quad |y - \hat{y}| \leq \delta \\ 
 \delta * (|y - \hat{y}|) - \frac{\delta^2}{2},   \quad otherwise 
 \end{cases} 
\end{align}
where $y$ is the actual value of the target variable and $\hat{y}$ is the predicted value of the target variable. 
%Huber is a combination of MAE and MSE loss functions trying to cover their cons. 
Essentially Huber loss combines the MSE and MAE loss functions to overcome their drawbacks. Huber loss function penalises with MSE for loss values smaller than $\delta$ which tries to minimise the average of errors. For larger errors when loss values are greater than $\delta$, Huber loss function penalises with a similar function to MAE and using the $\delta$ term as a regularizer to dampen the impact of the outliers.

%We then need to return the gradient and hessian into the training model, so the model can move towards the best values of parameters according to these functions. The gradient (G) and hessian (H) of the model was driven as follows: 
%\begin{align}
% & G= \frac{\partial{(c^Ty - c^T\hat{y})}}{\partial\hat{y}} = \frac{c^Ty}{\hat{y}^2}\\
% & H= \frac{\partial c^Ty}{\partial\hat{y}^2} = \frac{2x}{\hat{y}^3}
%\end{align}

%We replaced the above gradient and hessian in the objective function of the LightGBM model. The obtained results were not as good, so we do not report the results in this study. Our customised cost function is dominated by order values. That is, most of the output parameters that correspond to transshipment between hospitals take value equal to zero. However, the quantity orders from blood center take larger count values. This makes the output vector a large sparse vector that has larger elements in orders than transshipment. Therefore,  our customised loss function penalises any error in orders harshly than errors in transshipment making it less robust. A different set of outputs may work in favour of customised function. This needs to be further investigated by other data set and models. 

\section{Data and experimental setup}\label{sec:setupdata}

\subsection{Data and case study}\label{sec:dataset}
We gather data from a real-world case study for blood supply chain management that consists of four hospitals and a central blood bank. In such a network, hospitals can satisfy their demand by either ordering fresh blood units from the central blood bank or transshipping blood units from other hospitals in the network. These decisions depend on their demand and availability of blood in other hospitals. They have to make decisions at the beginning of each day when the demand for the day is still unknown. If they have excess blood units, a holding cost is incurred and if they have an old blood unit that is older than 11 days, they have to discard them with the cost of wastage.

Demand is uncertain and we assume it follows the zero-inflated negative binomial distribution. Zero-inflated negative binomial is a combination of negative binomial and logit distributions. For such a demand, the distribution outputs take non-negative integer values. This ought to be a realistic choice for demand as demand for blood units can be zero, for example over the weekends, or take a positive integer on weekdays. The distribution of zero-inflated negative binomial has three parameters: the number of trials ($r$) and the probability of success in each trial ($p$) that correspond to the negative binomial part of the distribution, and the inflated probability of zero ($\pi$) that corresponds to the logit part \citep{doyle2009examples}. We set these parameters for four hospitals as follows:  demand for hospital 1 takes ($\pi= 0.6 , r=4 , p=0.6$) , demand for hospital 2 takes  ($\pi= 0 . 6 , r = 3 , p = 0 . 57$) , demand for hospital 3 takes  ($ \pi= 0 . 25 , r = 15 , p = . 57$) and demand for hospital 4 takes  ($\pi= 0 . 25 , r = 15 , p = 0 . 48)$.

\subsection{Experimental setup}\label{sec:setup}

Our dataset for training ML models includes inventory level for four hospitals and the optimal values of orders for the entire course of the simulation. In total, we have 18,500 observations. This dataset is collected after running a 2SSP model to optimality (i.e., from an exact solver) over 18,500 days. The number of observations is set based on the simulation period that is chosen according to the Dvoretzky–Kiefer–Wolfowitz (DKF) inequality to ensure that at 95\% level of confidence the empirical distribution of total daily costs has an error less than 0.01 \citep{kosorok2008introduction}.
%We produce forecasts using three mentioned ML algorithms.
 
The input variables represent the inventory level of four hospitals, and the output variables correspond to the orders of each hospital from the blood centre (four decision variables), and transshipment quantities between one hospital and others with maximum shelf life of 11 days (132 decision variables = 4 hospitals $\times$ 11 blood units with different ages $\times$ 3 transshipment to three other hospitals). 
More specifically, the inputs of the ML models are the number of units with age $m (m=1, ..., M)$ at each hospital, i.e., the inventory level of each hospital, and the outputs are the orders of each hospital from the central blood bank and the transshipment of units with age $m$,$(m=1, ..., M)$ from hospital $i$ to $j$.
%

%We represent the inventory of each hospital as an 11 valued array where each element indicates the number of blood units of particular residual shelf life. Therefore, the input variable space of our modeling problem is 44 dimensional, each set of 11 elements represents the inventory levels of one of the hospitals. While the output space (decisions) includes 4 values of orders submitted to the central blood bank by each hospitals, and 132 (11 blood units with different ages × 4 origin hospitals ×3 other destination hospitals) values of transshipment of blood units with different ages between hospitals. 

%In order to solve such a problem, \citep{} proposed a two-stage stochastic programming model. 

We look at the utility of predictions (forecasts) to evaluate the performance of the ML models and the efficiency of the results. We measure their monetary value by replacing the predicted values of decision variables in the first stage of the objective function to compute the total costs of supply chain. The cost function is the total cost of the supply chain measured by inventory costs, ordering costs, transshipment costs, outdate, and shortage costs.

Note that once the optimal values of decision variables are predicted, we need to replace them in the supply chain to find out the values of other variables and their associated costs. We summarise this process as follows. We initiate the problem by providing the available inventory of blood units for each hospital. Then the ML models are used to predict the order quantities from the central bank as well as the transshipment quantities between hospitals. Once these values are determined, we calculate the orders and transshipment costs. Hospitals accordingly update their inventory level after transshipping the blood units.  Next, hospitals realise their demand and fulfil them using their available inventory. At this point since the demand is known, the corresponding shortage cost is calculated. The transshipped orders will be received at the end of the day and hospitals update their inventory level accordingly. At this stage, the outdate and holding costs are calculated. The total cost is simply the sum of the aforementioned costs.

%\begin{figure}[H]
%	\centering
%	\caption{Flowchart of the blood supply chain}
%	\includegraphics[scale=0.65]{simulation.pdf}
%	\label{fig:simulation_SCM}
%\end{figure}

\section{Empirical results and discussion}\label{sec:MLresults}

We implemented the ML methods as described in section \ref{sec:methodology} and looked at the forecast utility by evaluating the first stage of the objective function of the blood SC model (presented in \cite{DehghaniTwoStage}. That is, we first predicted the values of decision variables on daily basis (the number of blood units to be ordered from the central bank and transshipments between different hospitals). Then, we replaced the obtained values in the simulated blood supply chain model to obtain the transportation, holding, outdate, transshipment, and shortage costs over 18,500 days of simulation.  We benchmark our models against TS model, current policy and MLP model as the best performing model in \citep{abbasi2020predicting}.

Table \ref{table:results_s} displays the performance of the ML methods considered in this study as well as the TS model, current policy and MLP model. The results are reported for all involved costs including holding, transshipment, ordering, shortage, and outdate costs.

	\begin{table}[ht]
	\centering
	\caption{Forecasting performance of ML methods in terms of average imposed costs}
	\begin{tabular}{l|c|c|c|c|c|c}
		%		\toprule
		\hline
		\multicolumn{1}{l}{Models}&\multicolumn{1}{c}{Holding}&\multicolumn{1}{c}{Transshipment}&\multicolumn{1}{c}{Outdate}&\multicolumn{1}{c}{Ordering  }&\multicolumn{1}{c}{Shortage}&\multicolumn{1}{c}{Total}\\
		%		\midrule
		%\hline
		%\multicolumn{1}{l}{Models} & \multicolumn{1}{c}{Mean}  & \multicolumn{1}{c}{Mean} &  %  \multicolumn{1}{c}{Mean} &    \multicolumn{1}{c}{Mean} & %\multicolumn{1}{c}{Mean}\\
		\hline
		Ridge & 45.33    &  0.59  & 2.71  & 22.36    &  7.22   &  78.23 \\ 
    	SVR &  45.52  & 0.59  & 2.71 & \textbf{22.34}  & 7.47  & 78.49  \\ 
        LightGBM-MSE  & 46.19  &  \textbf{0.81}  & 3.15  &   22.40   &  7.49  & 80.06  \\ 

	    LightGBM-MAE  &45.47  & 1.77  & 1.88  & 22.45   &  5.02 &  76.54  \\ 
	    LightGBM-Huber   &47.60  & 1.36  & 1.88 &  22.45   & 5.13 &  78.43 \\  
	    MLP  & 51.13&   1.29   & 4.01  &22.61    & 5.91  & 84.95\\
	    TS model  & \textbf{42.73} &2.43     & \textbf{1.59}&    22.37 & 5.33 &\textbf{74.51}\\
	    Current policy  & 87.61& 3.61 &  2.51  &22.61&  \textbf{ 3.12} &   119.49\\
		\hline
	\end{tabular}\label{table:results_s}
\end{table}

The results based on the total cost indicate that, on average, LightGBM trained with MAE loss function (LightGBM-MAE) generates the lowest total costs in the supply chain. Further, the predicted results with LightGBM-MAE are very close to the optimal TS model, with LightGBM-MAE model performing only 2.6\% sub-optimal. Considering the other costs, we can see that the LightGBM-MAE model manages to outperform all other models, including the TS model, in reducing shortage costs. This is a great advantage for the model since blood shortages can have a dramatic impact in case of emergency. LightGBM-Huber results is somewhat between LightGBM-MAE and LightGBM-Huber as expected.

The current policy, on the other hand, has the smallest shortage costs at the expense of higher holding costs. The current policy has significantly higher holding costs than all other models which have contributed to a larger total cost. Observe that the current policy also has the largest transshipment cost. This indicates that the current policy is a very conservative approach that tries to minimise the shortage and outdate costs by transshipping the products between hospitals.

In terms of outdate costs, TS model has the best performance, followed by LightGBM-MAE and LightGBM-Huber models. SVR and Ridge models have the lowest transportation cost. This does not make them superior to other models because the lower transportation costs have contributed to higher shortage and outdate costs. As it is evident from the results, both Ridge and SVR models have a significantly higher shortage and higher outdate costs than other models except the MLP model that has the highest outdate cost. All proposed models outperform the MLP model which was the best performing model in the previous study carried out on the same data sets \citep{abbasi2020predicting}.

Comparing the performance of MLP and LightGBM-MAE models reveals that MLP has managed to perform well in ordering and consequently transshipment between different hospitals. However, MLP has performed poorly in terms of outdate and holding costs. Large holding and outdate costs verify that on one hand MLP model has not been able to predict decision variables accurately enough, thus resulted in higher holding and outdate costs. On the other hand, it misspecified the transshipments between hospitals. This, consequently, has caused relatively similar ordering costs to other models but higher holding and outdate costs.

The MLP model which was the outperforming model in \citep{abbasi2020predicting} utilised MSE as the loss function. More sophisticated type of Neural network models are capable of having customised loss function and has been shown promising for supply chain forecasting tasks \citep{salinas2020deepar}.

In summary, While LightGBM-MSE model manages to outperform the MLP model by 5\%, LightGBM-MAE predictions lead to about 10\% lower cost than MLP model.  The predicted results by LightGBM-MAE model are only 2.6\% more than the optimal TS model. This low cost is not only evident in the total cost, but LightGBM-MAE has consistently generated lower costs in holding, transshipment, outdate, ordering and shortage costs making it a robust model. The low cost of LightGBM-MAE model indicates that this model has effectively managed to predict demands and transship the blood units between hospitals while minimising the holding, outdate and shortage costs.

Our results show that the loss function can play a pivotal role in determining the performance of the models. This is even more apparent when the predictions are translated to monetary values. As it can be seen in Table \ref{table:results_s}, the LightGBM-MAE outperforms the counterpart LightGBM-MSE model by 5\%. However, different loss functions may perform better for a specific objective. For example, LightGBM-MSE has outperformed other models in terms of transshipment costs. While this may not be the optimal policy for the blood supply chain case, one can choose an appropriate model for minimising his desirable objective. We assert that it is imperative to choose the appropriate loss function according to the parameters of the model, the problem at hand and the interest of the decision-maker.

%For instance, if the cost of outdate is extremely large and we want to minimise that, then MAE can be an appropriate choice. In contrast, if the outdate, transportation, holding, and ordering are simialr, then MSE will result in better accuracy and lower costs.

\begin{figure}
	\centering
	\caption{The inventory level resulted by LGBM-MSE model}
	\includegraphics[scale=0.45]{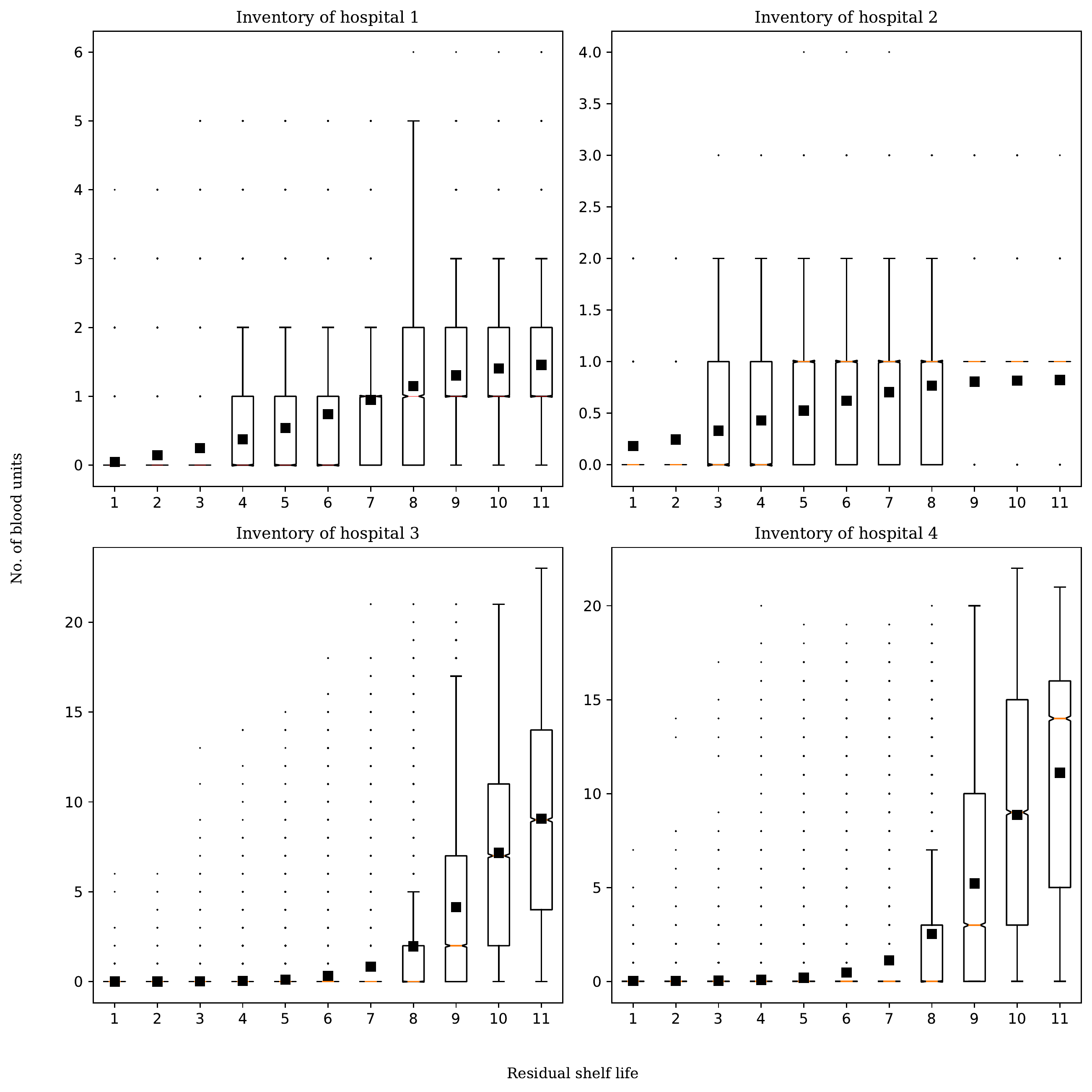}
	\label{fig:LGBMinventroy}
\end{figure}

\begin{figure}
	\centering
	\caption{The inventory level resulted by SVR model}
	\includegraphics[scale=0.45]{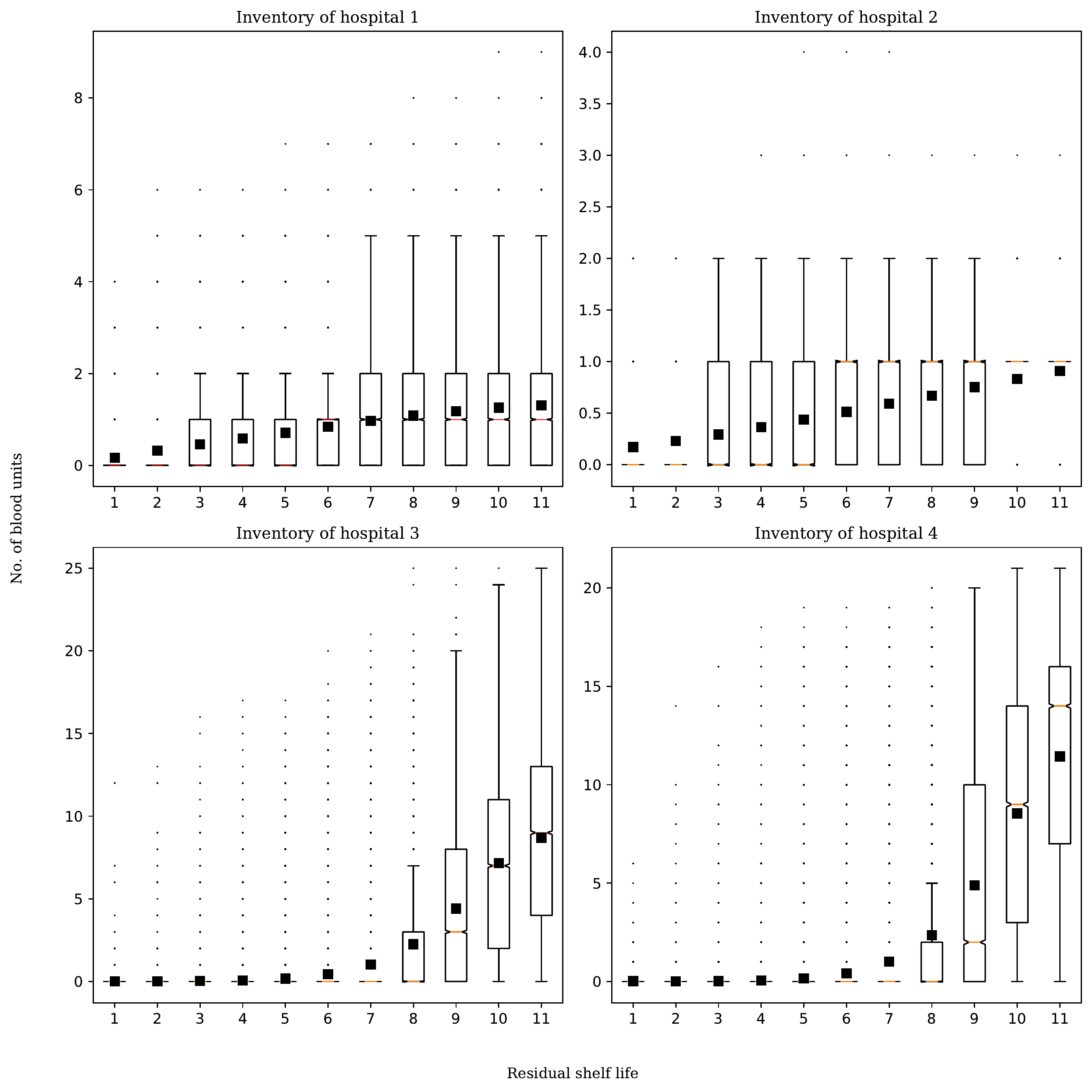}
	\label{fig:SVRinventroy}
\end{figure}

\begin{figure}
	\centering
	\caption{The inventory level resulted by Ridge model}
	\includegraphics[scale=0.45]{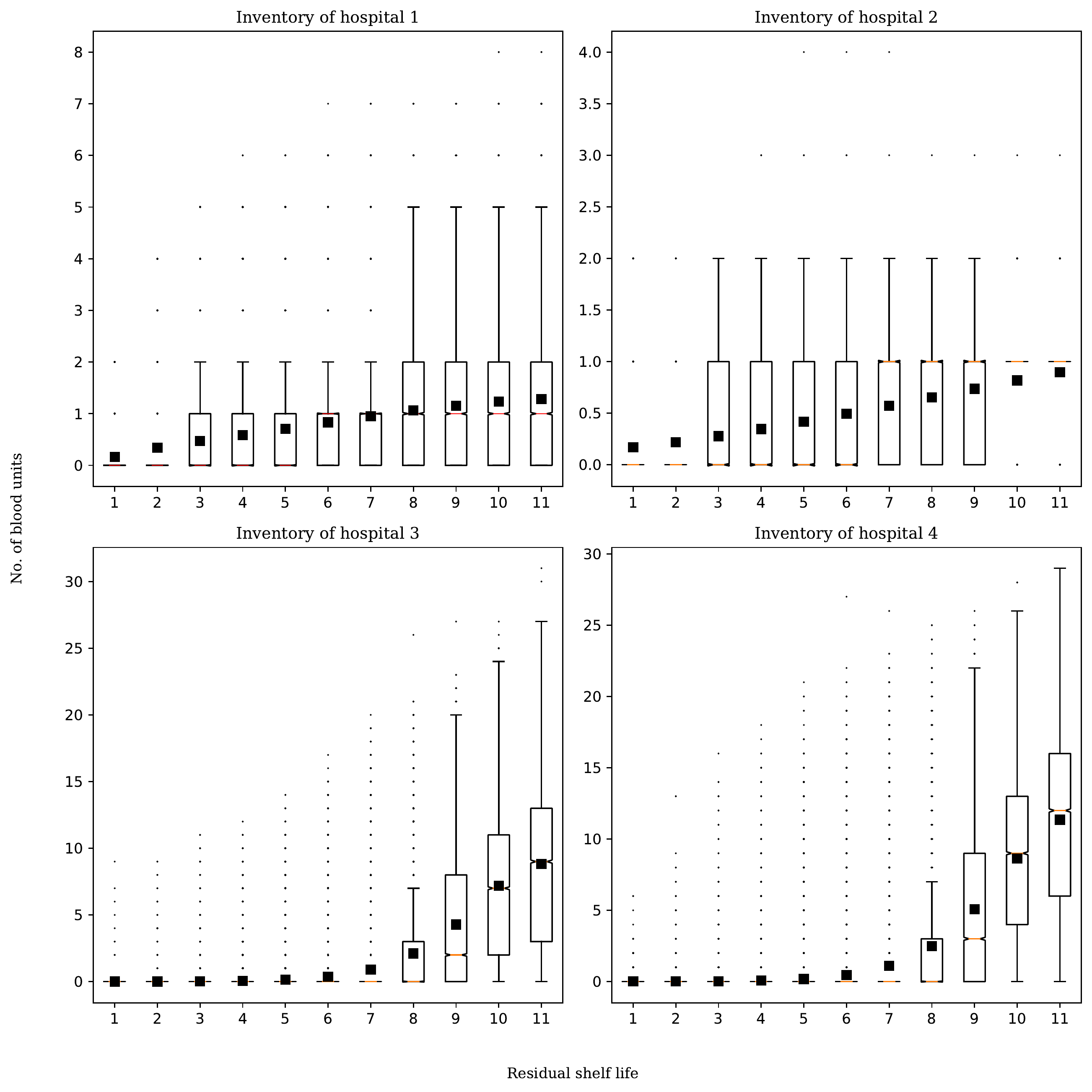}
	\label{fig:Ridgeinventroy}
\end{figure}

\begin{figure}
	\centering
	\caption{The inventory level resulted by LGBM-MAE model}
	\includegraphics[scale=0.45]{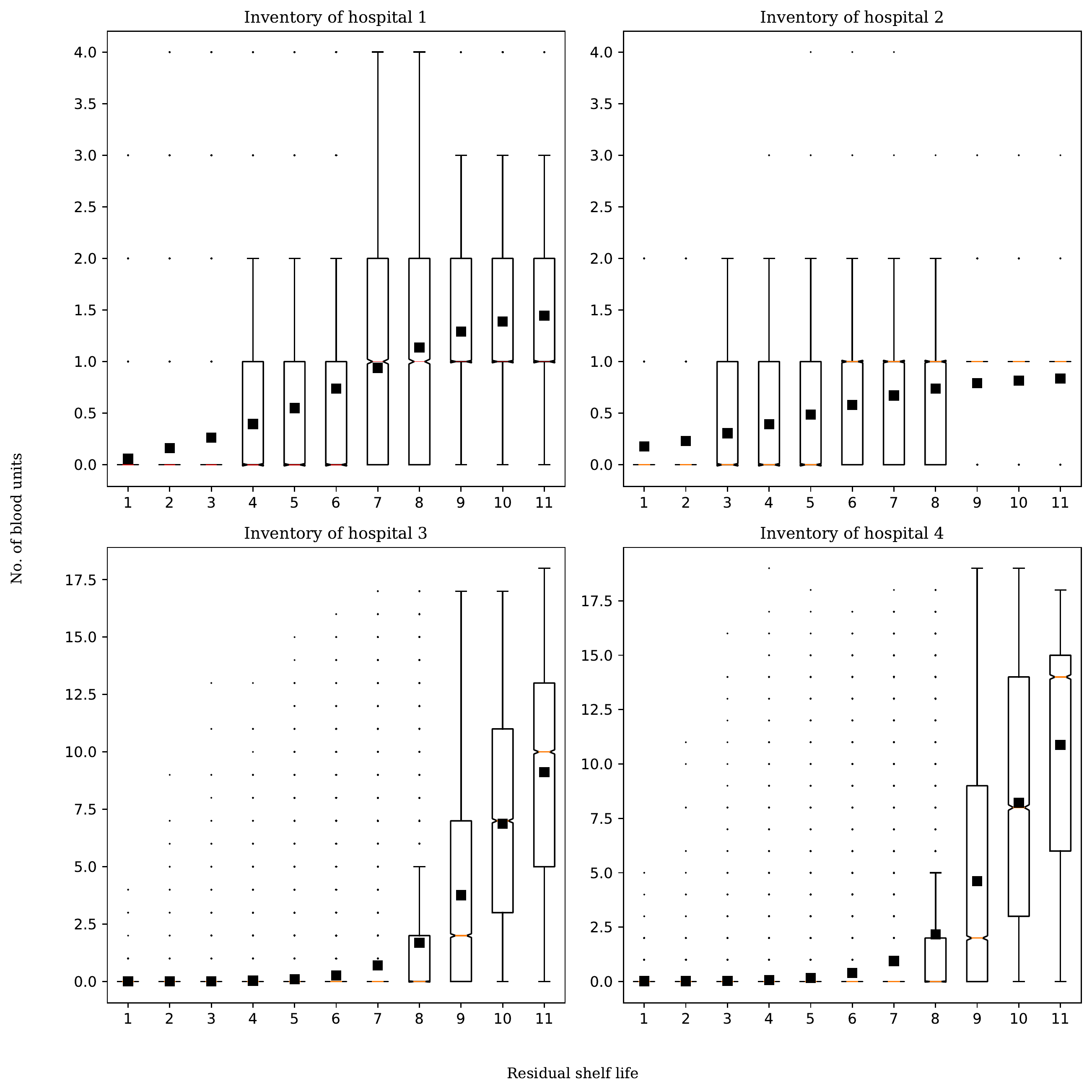}
	\label{fig:LGBM-MAEinventroy}
\end{figure}

\begin{figure}
	\centering
	\caption{The inventory level resulted by LGBM-Huber model}
	\includegraphics[scale=0.45]{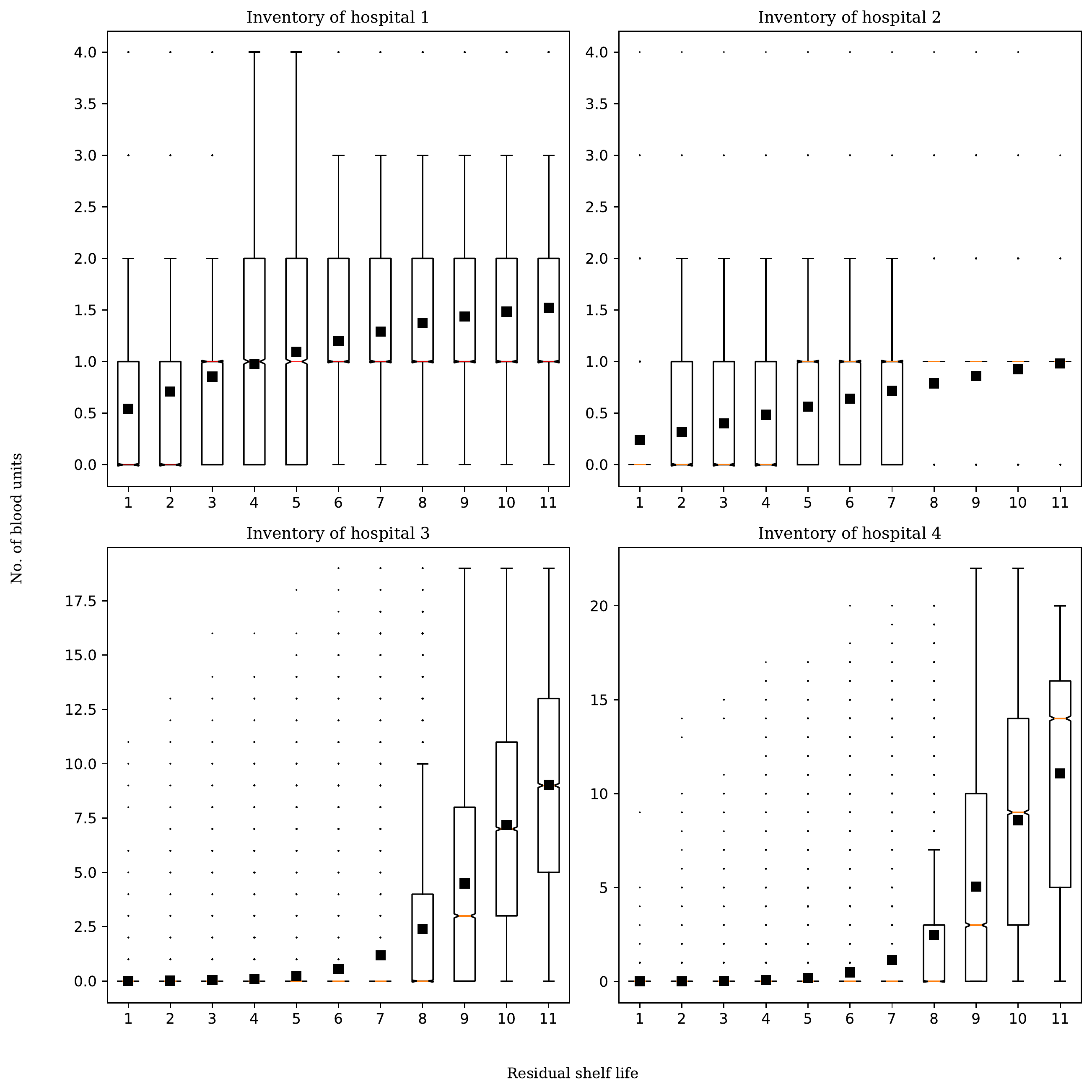}
	\label{fig:LGBM-Huberinventroy}
\end{figure}

In order to evaluate the performance of the models across all observations, we investigated the distribution of the various costs for Ridge, SVR, LightGBM-MAE, LightGBM-MSE, and LightGBM-Huber models at each hospital. The results, presented for each hospital and each ML model, are depicted in Figures \ref{fig:LGBMinventroy},\ref{fig:SVRinventroy}, \ref{fig:Ridgeinventroy}, \ref{fig:LGBM-MAEinventroy}, \ref{fig:LGBM-Huberinventroy} where box-plots are used to display the minimum, 1\textsuperscript{st} quantile, median, 3\textsuperscript{rd} quantile, and maximum values of the costs, as well as any possible outliers. As we can see from these graphs, the inventory levels for hospitals differ greatly for blood units with different ages. All hospitals have lower inventory levels for younger blood unis. Hospital 3 and 4 have the highest inventory levels for blood units with 9, 10, and 11 days old.  As it is evident from the distribution of inventory levels, the LightGBM-MAE model consistently generates a lower inventory level for all hospitals. SVR and Ridge models have generated a similar level of inventory for four different hospitals with Ridge model generating slightly lower levels of inventory.  We state that the LightGBM-MAE model is not only the most accurate model on average, it also generates the most accurate results across all hospitals and all days.

According to the obtained results in Table \ref{table:results_s}, we can assert that the MAE loss function is more appropriate for the blood supply chain problem and similar supply chain problems where we have a vector of output with many zero elements. LightGBM-MAE is particularly a good choice for modeling as it is flexible to change the loss function. 

As discussed before, when we use ML models to predict the solutions of constrained optimization problems, we should consider the constraints and whether the ML models commit to them or not. Table \ref{table:violations} shows how many times each of the presented ML models violated a constraint. That is, the ML model predicted a solution for the number of orders that was higher than the available inventory. 

	\begin{table}[ht]
	\centering
	\caption{The number of times ML models violated a constraint}
	\begin{tabular}{ l c c c c c }
		%		\toprule
		\hline
		 \multicolumn{1}{c}{Ridge}&\multicolumn{1}{c}{SVR}&\multicolumn{1}{c}{LightGBM-MSE}&\multicolumn{1}{c}{LightGBM-MAE  }&\multicolumn{1}{c}{LightGBM-Huber}&\multicolumn{1}{c}{MLP}\\
		\hline
		  209    &  21  & 71  & 21 &  58   &  411 \\ 
		\hline
	\end{tabular}\label{table:violations}
\end{table}

As it can be seen in Table \ref{table:violations}, the SVR and LightGBM-MAE models have a lower number of constraint violations with only 21 violations occurred among 814,000 order and inventory records. This indicates both of these models can mimic the solutions of constrained optimization models while significantly committing to the constraints. We can also see the impact of loss function as it has appeared in the number violated constraints. We observe that the generated solutions from an MSE loss function have violated the constraints more frequently than other loss functions, making it less attractive for such a predicting task. Lastly, we can see that MLP has performed poorly in comparison to other investigated models. 
%Loss function is an important part of the forecasting problem. While in academia statistical loss functions are widely used for training and evaluating models performances, loss function is often in dollar terms for managers. The true loss function is often difficult to estimate as its distribution is unknown. There is no consensus in the literature in choosing the best loss function \citep{clements1995selection,clements1993limitations}, rather it depends on the dataset, problem at hand , and the objectives of decision maker. Reliability, robustness to outliers and comprehensibility are some of the desirable criteria for good loss function. 

\section{Conclusion}\label{sec:MLconclusion}

In this paper, we introduced a framework to use multi-output ML models that are trained according to different cost functions to predict the solutions for large-scale constrained optimization models. Our results indicate that we can use ML models to forecast the optimal values of the parameters with up to 98\% similarity to the optimal solution while committing to the constraints over 99\% of the times. We investigated the role of loss function in predicting the solutions for optimization problems. To do so, we trained LightGBM models with MAE, MSE and Huber loss functions and showed that using MAE loss function often leads to a better performance than using MSE loss function, while Huber loss averages the MAE and MSE results. 
Therefore, we suggest using  ML models with appropriate loss functions to predict the solutions of optimizations models. While well-tuned ML models can generate competitive results, they perform significantly faster than the commercial solvers and they are cheaper in terms of costs.

We further explored the performance of the ML models by investigating the utility of forecasts and examining different costs associated with the generated solutions. We computed the holding, transshipment, outdating, ordering and shortage costs and showed that different models perform differently and a model should be chosen based on the desired criteria. 

In this study, we focused on the total daily costs of the supply chain and optimized our algorithm to minimise the total cost. However, one can consider different objectives to optimize the learning process of the ML model.
One natural extension to this research is to train the ML models with a customised loss function. This customised loss function can be the objective function of the problem at hand. In order to replace the objective function with a customised function, we need to derive the gradient and hessian of the objective function which can potentially improve the learning process and lead to better performance. 

More studies are required to test whether our results apply to other supply chain problems. A similar finding by other researchers can enormously benefit the operations research society and practitioners by using free and fast ML models to solve constrained optimization problems. There is a need for research to better understand the impact of loss function on the accuracy of models that are trained for mimicking constrained optimization problems. 
%Other researchers can be focused at solving such ML supply chain problems by multi-label classification models. 

%\section*{Acknowledgment}
%The authors are grateful to the review team and the area editor for providing constructive suggestions and comments.

%\section*{References}
\bibliography{biblio}

\begin{thebibliography}{41}
\expandafter\ifx\csname natexlab\endcsname\relax\def\natexlab#1{#1}\fi
\providecommand{\url}[1]{\texttt{#1}}
\providecommand{\href}[2]{#2}
\providecommand{\path}[1]{#1}
\providecommand{\DOIprefix}{doi:}
\providecommand{\ArXivprefix}{arXiv:}
\providecommand{\URLprefix}{URL: }
\providecommand{\Pubmedprefix}{pmid:}
\providecommand{\doi}[1]{\href{http://dx.doi.org/#1}{\path{#1}}}
\providecommand{\Pubmed}[1]{\href{pmid:#1}{\path{#1}}}
\providecommand{\bibinfo}[2]{#2}
\ifx\xfnm\relax \def\xfnm[#1]{\unskip,\space#1}\fi
%Type = Article
\bibitem[{Abbasi et~al.(2020)Abbasi, Babaei, Hosseinifard, Smith-Miles and
  Dehghani}]{abbasi2020predicting}
\bibinfo{author}{Abbasi, B.}, \bibinfo{author}{Babaei, T.},
  \bibinfo{author}{Hosseinifard, Z.}, \bibinfo{author}{Smith-Miles, K.},
  \bibinfo{author}{Dehghani, M.}, \bibinfo{year}{2020}.
\newblock \bibinfo{title}{Predicting solutions of large-scale optimization
  problems via machine learning: A case study in blood supply chain
  management}.
\newblock \bibinfo{journal}{Computers \& Operations Research}
  \bibinfo{volume}{119}, \bibinfo{pages}{104941}.
%Type = Article
\bibitem[{Abolghasemi et~al.(2020a)Abolghasemi, Beh, Tarr and
  Gerlach}]{abolghasemi2020demand}
\bibinfo{author}{Abolghasemi, M.}, \bibinfo{author}{Beh, E.},
  \bibinfo{author}{Tarr, G.}, \bibinfo{author}{Gerlach, R.},
  \bibinfo{year}{2020}a.
\newblock \bibinfo{title}{Demand forecasting in supply chain: The impact of
  demand volatility in the presence of promotion}.
\newblock \bibinfo{journal}{Computers \& Industrial Engineering} ,
  \bibinfo{pages}{106380}.
%Type = Article
\bibitem[{Abolghasemi et~al.(2020b)Abolghasemi, Hurley, Eshragh and
  Fahimnia}]{abolghasemi2020demand2}
\bibinfo{author}{Abolghasemi, M.}, \bibinfo{author}{Hurley, J.},
  \bibinfo{author}{Eshragh, A.}, \bibinfo{author}{Fahimnia, B.},
  \bibinfo{year}{2020}b.
\newblock \bibinfo{title}{Demand forecasting in the presence of systematic
  events: Cases in capturing sales promotions}.
\newblock \bibinfo{journal}{International Journal of Production Economics} ,
  \bibinfo{pages}{107892}.
%Type = Article
\bibitem[{Abolghasemi et~al.(2019)Abolghasemi, Hyndman, Tarr and
  Bergmeir}]{abolghasemi2019machine}
\bibinfo{author}{Abolghasemi, M.}, \bibinfo{author}{Hyndman, R.J.},
  \bibinfo{author}{Tarr, G.}, \bibinfo{author}{Bergmeir, C.},
  \bibinfo{year}{2019}.
\newblock \bibinfo{title}{Machine learning applications in time series
  hierarchical forecasting}.
\newblock \bibinfo{journal}{arXiv preprint arXiv:1912.00370} .
%Type = Article
\bibitem[{Ali et~al.(2012)Ali, Boylan and Syntetos}]{ali2012forecast}
\bibinfo{author}{Ali, M.M.}, \bibinfo{author}{Boylan, J.E.},
  \bibinfo{author}{Syntetos, A.A.}, \bibinfo{year}{2012}.
\newblock \bibinfo{title}{Forecast errors and inventory performance under
  forecast information sharing}.
\newblock \bibinfo{journal}{International Journal of Forecasting}
  \bibinfo{volume}{28}, \bibinfo{pages}{830--841}.
%Type = Book
\bibitem[{Armstrong(2001)}]{armstrong2001principles}
\bibinfo{author}{Armstrong, J.S.}, \bibinfo{year}{2001}.
\newblock \bibinfo{title}{Principles of forecasting: a handbook for researchers
  and practitioners}. volume~\bibinfo{volume}{30}.
\newblock \bibinfo{publisher}{Springer Science \& Business Media}.
%Type = Inproceedings
\bibitem[{Barron(2019)}]{barron2019general}
\bibinfo{author}{Barron, J.T.}, \bibinfo{year}{2019}.
\newblock \bibinfo{title}{A general and adaptive robust loss function}, in:
  \bibinfo{booktitle}{Proceedings of the IEEE/CVF Conference on Computer Vision
  and Pattern Recognition}, pp. \bibinfo{pages}{4331--4339}.
%Type = Article
\bibitem[{Bengio et~al.(2020)Bengio, Lodi and Prouvost}]{bengio2020MLsurvey}
\bibinfo{author}{Bengio, Y.}, \bibinfo{author}{Lodi, A.},
  \bibinfo{author}{Prouvost, A.}, \bibinfo{year}{2020}.
\newblock \bibinfo{title}{Machine learning for combinatorial optimization: A
  methodological tour d’horizon}.
\newblock \bibinfo{journal}{Management Science} \bibinfo{volume}{290},
  \bibinfo{pages}{405 -- 421}.
%Type = Article
\bibitem[{Bojer and Meldgaard(2020)}]{bojer2020kaggle}
\bibinfo{author}{Bojer, C.S.}, \bibinfo{author}{Meldgaard, J.P.},
  \bibinfo{year}{2020}.
\newblock \bibinfo{title}{Kaggle forecasting competitions: An overlooked
  learning opportunity}.
\newblock \bibinfo{journal}{International Journal of Forecasting} .
%Type = Article
\bibitem[{Boylan et~al.(2006)Boylan, Syntetos et~al.}]{boylan2006accuracy}
\bibinfo{author}{Boylan, J.E.}, \bibinfo{author}{Syntetos, A.A.}, et~al.,
  \bibinfo{year}{2006}.
\newblock \bibinfo{title}{Accuracy and accuracy-implication metrics for
  intermittent demand}.
\newblock \bibinfo{journal}{Foresight: The International Journal of Applied
  Forecasting} \bibinfo{volume}{4}, \bibinfo{pages}{39--42}.
%Type = Article
\bibitem[{Burges(1998)}]{burges1998tutorial}
\bibinfo{author}{Burges, C.J.}, \bibinfo{year}{1998}.
\newblock \bibinfo{title}{A tutorial on support vector machines for pattern
  recognition}.
\newblock \bibinfo{journal}{Data mining and knowledge discovery}
  \bibinfo{volume}{2}, \bibinfo{pages}{121--167}.
%Type = Article
\bibitem[{Christoffersen and Diebold(1997)}]{christoffersen1997optimal}
\bibinfo{author}{Christoffersen, P.F.}, \bibinfo{author}{Diebold, F.X.},
  \bibinfo{year}{1997}.
\newblock \bibinfo{title}{Optimal prediction under asymmetric loss}.
\newblock \bibinfo{journal}{Econometric theory} , \bibinfo{pages}{808--817}.
%Type = Article
\bibitem[{Clements and Hendry(1995)}]{clements1995selection}
\bibinfo{author}{Clements, M.}, \bibinfo{author}{Hendry, D.},
  \bibinfo{year}{1995}.
\newblock \bibinfo{title}{On the selection of error measures for comparisons
  among forecasting methods-reply}.
\newblock \bibinfo{journal}{Journal of Forecasting} \bibinfo{volume}{14},
  \bibinfo{pages}{73--75}.
%Type = Article
\bibitem[{Clements and Hendry(1993)}]{clements1993limitations}
\bibinfo{author}{Clements, M.P.}, \bibinfo{author}{Hendry, D.F.},
  \bibinfo{year}{1993}.
\newblock \bibinfo{title}{On the limitations of comparing mean square forecast
  errors}.
\newblock \bibinfo{journal}{Journal of Forecasting} \bibinfo{volume}{12},
  \bibinfo{pages}{617--637}.
%Type = Article
\bibitem[{Dantzig(1955)}]{dantzig1955SSP}
\bibinfo{author}{Dantzig, G.B.}, \bibinfo{year}{1955}.
\newblock \bibinfo{title}{Linear programming under uncertainty}.
\newblock \bibinfo{journal}{Management Science} \bibinfo{volume}{1},
  \bibinfo{pages}{197 -- 206}.
%Type = Article
\bibitem[{Dehghani and Abbasi(2020)}]{Dehghani2018}
\bibinfo{author}{Dehghani, M.}, \bibinfo{author}{Abbasi, B.},
  \bibinfo{year}{2020}.
\newblock \bibinfo{title}{An age-based lateral-transshipment policy for
  perishable items}.
\newblock \bibinfo{journal}{Production Economics} \bibinfo{volume}{198},
  \bibinfo{pages}{93 -- 103}.
%Type = Article
\bibitem[{Dehghani et~al.(2019)Dehghani, Abbasi and
  Oliveira}]{DehghaniTwoStage}
\bibinfo{author}{Dehghani, M.}, \bibinfo{author}{Abbasi, B.},
  \bibinfo{author}{Oliveira, F.}, \bibinfo{year}{2019}.
\newblock \bibinfo{title}{Proactive transshipment in the blood supply chain: a
  stochastic programming approach}.
\newblock \bibinfo{journal}{Omega} \bibinfo{volume}{Online-published},
  \bibinfo{pages}{1--16}.
%Type = Article
\bibitem[{Doyle(2009)}]{doyle2009examples}
\bibinfo{author}{Doyle, S.R.}, \bibinfo{year}{2009}.
\newblock \bibinfo{title}{Examples of computing power for zero-inflated and
  overdispersed count data}.
\newblock \bibinfo{journal}{Journal of Modern Applied Statistical Methods}
  \bibinfo{volume}{8}, \bibinfo{pages}{3}.
%Type = Article
\bibitem[{Drucker et~al.(1997)Drucker, Burges, Kaufman, Smola, Vapnik
  et~al.}]{drucker1997support}
\bibinfo{author}{Drucker, H.}, \bibinfo{author}{Burges, C.J.},
  \bibinfo{author}{Kaufman, L.}, \bibinfo{author}{Smola, A.},
  \bibinfo{author}{Vapnik, V.}, et~al., \bibinfo{year}{1997}.
\newblock \bibinfo{title}{Support vector regression machines}.
\newblock \bibinfo{journal}{Advances in neural information processing systems}
  \bibinfo{volume}{9}, \bibinfo{pages}{155--161}.
%Type = Article
\bibitem[{Exterkate et~al.(2016)Exterkate, Groenen, Heij and van
  Dijk}]{exterkate2016nonlinear}
\bibinfo{author}{Exterkate, P.}, \bibinfo{author}{Groenen, P.J.},
  \bibinfo{author}{Heij, C.}, \bibinfo{author}{van Dijk, D.},
  \bibinfo{year}{2016}.
\newblock \bibinfo{title}{Nonlinear forecasting with many predictors using
  kernel ridge regression}.
\newblock \bibinfo{journal}{International Journal of Forecasting}
  \bibinfo{volume}{32}, \bibinfo{pages}{736--753}.
%Type = Article
\bibitem[{Fildes and Petropoulos(2015)}]{FILDES20151692}
\bibinfo{author}{Fildes, R.}, \bibinfo{author}{Petropoulos, F.},
  \bibinfo{year}{2015}.
\newblock \bibinfo{title}{{Simple versus complex selection rules for
  forecasting many time series}}.
\newblock \bibinfo{journal}{Journal of Business Research} \bibinfo{volume}{68},
  \bibinfo{pages}{1692--1701}.
\newblock \DOIprefix\doi{https://doi.org/10.1016/j.jbusres.2015.03.028}.
  \bibinfo{note}{special Issue on Simple Versus Complex Forecasting}.
%Type = Article
\bibitem[{Hoerl and Kennard(1970)}]{hoerl1970ridge}
\bibinfo{author}{Hoerl, A.E.}, \bibinfo{author}{Kennard, R.W.},
  \bibinfo{year}{1970}.
\newblock \bibinfo{title}{Ridge regression: Biased estimation for nonorthogonal
  problems}.
\newblock \bibinfo{journal}{Technometrics} \bibinfo{volume}{12},
  \bibinfo{pages}{55--67}.
%Type = Article
\bibitem[{Ke et~al.(2017)Ke, Meng, Finley, Wang, Chen, Ma, Ye and
  Liu}]{ke2017lightgbm}
\bibinfo{author}{Ke, G.}, \bibinfo{author}{Meng, Q.}, \bibinfo{author}{Finley,
  T.}, \bibinfo{author}{Wang, T.}, \bibinfo{author}{Chen, W.},
  \bibinfo{author}{Ma, W.}, \bibinfo{author}{Ye, Q.}, \bibinfo{author}{Liu,
  T.Y.}, \bibinfo{year}{2017}.
\newblock \bibinfo{title}{Lightgbm: A highly efficient gradient boosting
  decision tree}.
\newblock \bibinfo{journal}{Advances in neural information processing systems}
  \bibinfo{volume}{30}, \bibinfo{pages}{3146--3154}.
%Type = Book
\bibitem[{Kosorok(2008)}]{kosorok2008introduction}
\bibinfo{author}{Kosorok, M.R.}, \bibinfo{year}{2008}.
\newblock \bibinfo{title}{Introduction to empirical processes and
  semiparametric inference.}
\newblock \bibinfo{publisher}{Springer}.
%Type = Inproceedings
\bibitem[{Kruber et~al.(2017)Kruber, L{\"u}bbecke and
  Parmentier}]{kruber2017learning}
\bibinfo{author}{Kruber, M.}, \bibinfo{author}{L{\"u}bbecke, M.E.},
  \bibinfo{author}{Parmentier, A.}, \bibinfo{year}{2017}.
\newblock \bibinfo{title}{Learning when to use a decomposition}, in:
  \bibinfo{booktitle}{International Conference on AI and OR Techniques in
  Constraint Programming for Combinatorial Optimization Problems},
  \bibinfo{organization}{Springer}. pp. \bibinfo{pages}{202--210}.
%Type = Article
\bibitem[{Larsen et~al.(2021)Larsen, Lachapelle, Bengio, Frejinger,
  Lacoste-Julien and Lodi}]{larsen2021predicting}
\bibinfo{author}{Larsen, E.}, \bibinfo{author}{Lachapelle, S.},
  \bibinfo{author}{Bengio, Y.}, \bibinfo{author}{Frejinger, E.},
  \bibinfo{author}{Lacoste-Julien, S.}, \bibinfo{author}{Lodi, A.},
  \bibinfo{year}{2021}.
\newblock \bibinfo{title}{Predicting tactical solutions to operational planning
  problems under imperfect information}
  \href{http://arxiv.org/abs/1807.11876}{{\tt arXiv:1807.11876}}.
%Type = Article
\bibitem[{Levis and Papageorgiou(2005)}]{levis2005customer}
\bibinfo{author}{Levis, A.}, \bibinfo{author}{Papageorgiou, L.},
  \bibinfo{year}{2005}.
\newblock \bibinfo{title}{Customer demand forecasting via support vector
  regression analysis}.
\newblock \bibinfo{journal}{Chemical Engineering Research and Design}
  \bibinfo{volume}{83}, \bibinfo{pages}{1009--1018}.
%Type = Inproceedings
\bibitem[{Lin et~al.(2017)Lin, Goyal, Girshick, He and
  Doll{\'a}r}]{lin2017focal}
\bibinfo{author}{Lin, T.Y.}, \bibinfo{author}{Goyal, P.},
  \bibinfo{author}{Girshick, R.}, \bibinfo{author}{He, K.},
  \bibinfo{author}{Doll{\'a}r, P.}, \bibinfo{year}{2017}.
\newblock \bibinfo{title}{Focal loss for dense object detection}, in:
  \bibinfo{booktitle}{Proceedings of the IEEE international conference on
  computer vision}, pp. \bibinfo{pages}{2980--2988}.
%Type = Article
\bibitem[{Lodi and Zarpellon(2017)}]{lodi2017learning}
\bibinfo{author}{Lodi, A.}, \bibinfo{author}{Zarpellon, G.},
  \bibinfo{year}{2017}.
\newblock \bibinfo{title}{On learning and branching: a survey}.
\newblock \bibinfo{journal}{TOP} \bibinfo{volume}{25},
  \bibinfo{pages}{207--236}.
%Type = Article
\bibitem[{Makridakis and Hibon(1991)}]{makridakis1991exponential}
\bibinfo{author}{Makridakis, S.}, \bibinfo{author}{Hibon, M.},
  \bibinfo{year}{1991}.
\newblock \bibinfo{title}{Exponential smoothing: The effect of initial values
  and loss functions on post-sample forecasting accuracy}.
\newblock \bibinfo{journal}{International Journal of Forecasting}
  \bibinfo{volume}{7}, \bibinfo{pages}{317--330}.
%Type = Article
\bibitem[{Makridakis and Spiliotis(2021)}]{makridakis2021m5}
\bibinfo{author}{Makridakis, S.}, \bibinfo{author}{Spiliotis, E.},
  \bibinfo{year}{2021}.
\newblock \bibinfo{title}{The m5 competition and the future of human expertise
  in forecasting.}
\newblock \bibinfo{journal}{Foresight: The International Journal of Applied
  Forecasting} .
%Type = Article
\bibitem[{Marcos~Alvarez et~al.(2016)Marcos~Alvarez, Wehenkel and
  Louveaux}]{marcos2016online}
\bibinfo{author}{Marcos~Alvarez, A.}, \bibinfo{author}{Wehenkel, L.},
  \bibinfo{author}{Louveaux, Q.}, \bibinfo{year}{2016}.
\newblock \bibinfo{title}{Online learning for strong branching approximation in
  branch-and-bound} .
%Type = Article
\bibitem[{Montero-Manso et~al.(2020)Montero-Manso, Athanasopoulos, Hyndman and
  Talagala}]{MONTEROMANSO202086}
\bibinfo{author}{Montero-Manso, P.}, \bibinfo{author}{Athanasopoulos, G.},
  \bibinfo{author}{Hyndman, R.J.}, \bibinfo{author}{Talagala, T.S.},
  \bibinfo{year}{2020}.
\newblock \bibinfo{title}{Fforma: Feature-based forecast model averaging}.
\newblock \bibinfo{journal}{International Journal of Forecasting}
  \bibinfo{volume}{36}, \bibinfo{pages}{86--92}.
\newblock \DOIprefix\doi{https://doi.org/10.1016/j.ijforecast.2019.02.011}.
%Type = Article
\bibitem[{Mossina et~al.(2019)Mossina, Rachelson and
  Delahaye}]{mossina2019multi}
\bibinfo{author}{Mossina, L.}, \bibinfo{author}{Rachelson, E.},
  \bibinfo{author}{Delahaye, D.}, \bibinfo{year}{2019}.
\newblock \bibinfo{title}{Multi-label classification for the generation of
  sub-problems in time-constrained combinatorial optimization} ,
  \bibinfo{pages}{1--9}.
%Type = Article
\bibitem[{Petropoulos et~al.(2014)Petropoulos, Makridakis, Assimakopoulos and
  Nikolopoulos}]{Petropoulos2014a}
\bibinfo{author}{Petropoulos, F.}, \bibinfo{author}{Makridakis, S.},
  \bibinfo{author}{Assimakopoulos, V.}, \bibinfo{author}{Nikolopoulos, K.},
  \bibinfo{year}{2014}.
\newblock \bibinfo{title}{{'Horses for Courses' in demand forecasting}}.
\newblock \bibinfo{journal}{European Journal of Operational Research}
  \bibinfo{volume}{237}, \bibinfo{pages}{152--163}.
%Type = Article
\bibitem[{Salinas et~al.(2020)Salinas, Flunkert, Gasthaus and
  Januschowski}]{salinas2020deepar}
\bibinfo{author}{Salinas, D.}, \bibinfo{author}{Flunkert, V.},
  \bibinfo{author}{Gasthaus, J.}, \bibinfo{author}{Januschowski, T.},
  \bibinfo{year}{2020}.
\newblock \bibinfo{title}{Deepar: Probabilistic forecasting with autoregressive
  recurrent networks}.
\newblock \bibinfo{journal}{International Journal of Forecasting}
  \bibinfo{volume}{36}, \bibinfo{pages}{1181--1191}.
%Type = Article
\bibitem[{Smith-Miles(2009)}]{smith2009cross}
\bibinfo{author}{Smith-Miles, K.A.}, \bibinfo{year}{2009}.
\newblock \bibinfo{title}{Cross-disciplinary perspectives on meta-learning for
  algorithm selection}.
\newblock \bibinfo{journal}{ACM Computing Surveys (CSUR)} \bibinfo{volume}{41},
  \bibinfo{pages}{1--25}.
%Type = Article
\bibitem[{Syntetos et~al.(2016)Syntetos, Babai, Boylan, Kolassa and
  Nikolopoulos}]{syntetos2016supply}
\bibinfo{author}{Syntetos, A.A.}, \bibinfo{author}{Babai, Z.},
  \bibinfo{author}{Boylan, J.E.}, \bibinfo{author}{Kolassa, S.},
  \bibinfo{author}{Nikolopoulos, K.}, \bibinfo{year}{2016}.
\newblock \bibinfo{title}{Supply chain forecasting: Theory, practice, their gap
  and the future}.
\newblock \bibinfo{journal}{European Journal of Operational Research}
  \bibinfo{volume}{252}, \bibinfo{pages}{1--26}.
%Type = Article
\bibitem[{Syntetos and Boylan(2006)}]{syntetos2006stock}
\bibinfo{author}{Syntetos, A.A.}, \bibinfo{author}{Boylan, J.E.},
  \bibinfo{year}{2006}.
\newblock \bibinfo{title}{On the stock control performance of intermittent
  demand estimators}.
\newblock \bibinfo{journal}{International Journal of Production Economics}
  \bibinfo{volume}{103}, \bibinfo{pages}{36--47}.
%Type = Article
\bibitem[{Vaclavik et~al.(2018)Vaclavik, Novak, Scha and
  Hanzlek}]{vaclavik2018accelerating}
\bibinfo{author}{Vaclavik, R.}, \bibinfo{author}{Novak, A.},
  \bibinfo{author}{Scha, P.}, \bibinfo{author}{Hanzlek, Z.},
  \bibinfo{year}{2018}.
\newblock \bibinfo{title}{Accelerating the branch-and-price algorithm using
  machine learning}.
\newblock \bibinfo{journal}{European Journal of Operational Research} .
%Type = Article
\bibitem[{Zamani et~al.(2020)Zamani, Abolghasemi, Hosseini and
  Pishvaee}]{zamani2020considering}
\bibinfo{author}{Zamani, M.}, \bibinfo{author}{Abolghasemi, M.},
  \bibinfo{author}{Hosseini, S.M.S.}, \bibinfo{author}{Pishvaee, M.S.},
  \bibinfo{year}{2020}.
\newblock \bibinfo{title}{Considering pricing and uncertainty in designing a
  reverse logistics network}.
\newblock \bibinfo{journal}{International Journal of Industrial and Systems
  Engineering} \bibinfo{volume}{35}, \bibinfo{pages}{158--182}.

\end{thebibliography}

\end{document}